\documentclass[11pt,a4paper]{article}
\usepackage[hyperref]{emnlp-ijcnlp-2019}
\usepackage{times}
\usepackage{latexsym}
\usepackage{amsmath}
\usepackage{mathrsfs}
\usepackage{url}
\usepackage{graphicx}
\usepackage{epstopdf}
\usepackage{multirow}
\usepackage{enumitem}
\usepackage{amsfonts}
\usepackage{tabularx,booktabs}
\usepackage[font=small,labelfont=bf,textfont=md]{caption}
\usepackage{float}
\usepackage{url}

\newcommand{\smallsection}[1]{\smallskip \noindent\textbf{#1.}}

\newcolumntype{Y}{>{\centering\arraybackslash}X}
\newcolumntype{L}{>{\arraybackslash}X}

\def \h {\mathbf{h}}
\def \r {\mathbf{r}}
\def \D {\mathcal{D}}
\def \F {\mathcal{F}}
\def \Y {Y}
\def \Trans {^{T}}
\def \eg {\textit{e.g.}}
\def \ie {\textit{i.e.}}

\aclfinalcopy 

\author{Qinyuan Ye\thanks{\; Equal contribution.} \\
  University of Southern California\\
  \texttt{qinyuany@usc.edu} \\\And
  Liyuan Liu$^*$ \\
  University of Illinois, Urbana-Champaign\\
  \texttt{ll2@illinois.edu} \\\AND
   Maosen Zhang\\
  Purdue University \\
  \texttt{maosenzhang.milo@gmail.com} \\\And
  Xiang Ren \\
  University of Southern California\\
  \texttt{xiangren@usc.edu}\\
  }

\date{}

\begin{document}

\title{Looking Beyond Label Noise: Shifted Label Distribution Matters \\in Distantly Supervised Relation Extraction}
\maketitle


\begin{abstract}
In recent years there is a surge of interest in applying distant supervision (DS) to automatically generate training data for relation extraction (RE).
In this paper, we study the problem \textit{what limits the performance of DS-trained neural models}, conduct thorough analyses, and identify a factor that can influence the performance greatly, \textit{shifted label distribution}. 
Specifically, we found 
this problem commonly exists in real-world DS datasets, 
and without special handing, typical DS-RE models cannot automatically adapt to this shift, thus achieving deteriorated performance. 
To further validate our intuition, we develop a simple yet effective adaptation method for DS-trained models, \textit{bias adjustment}, which updates models learned over the source domain (\ie, DS training set) with a label distribution estimated on the target domain (\ie, test set).
Experiments demonstrate that bias adjustment achieves consistent performance gains on DS-trained models, especially on neural models, with an up to 23\% relative F1 improvement, which verifies our assumptions. 
Our code and data can be found at  \url{https://github.com/INK-USC/shifted-label-distribution}.
\end{abstract}

\section{Introduction}
\begin{figure}
\centering
\begin{minipage}{.29\textwidth}
  \centering
  \includegraphics[trim={2.1cm, 1.2cm, 0.5cm, 1.5cm},clip,width=\linewidth]{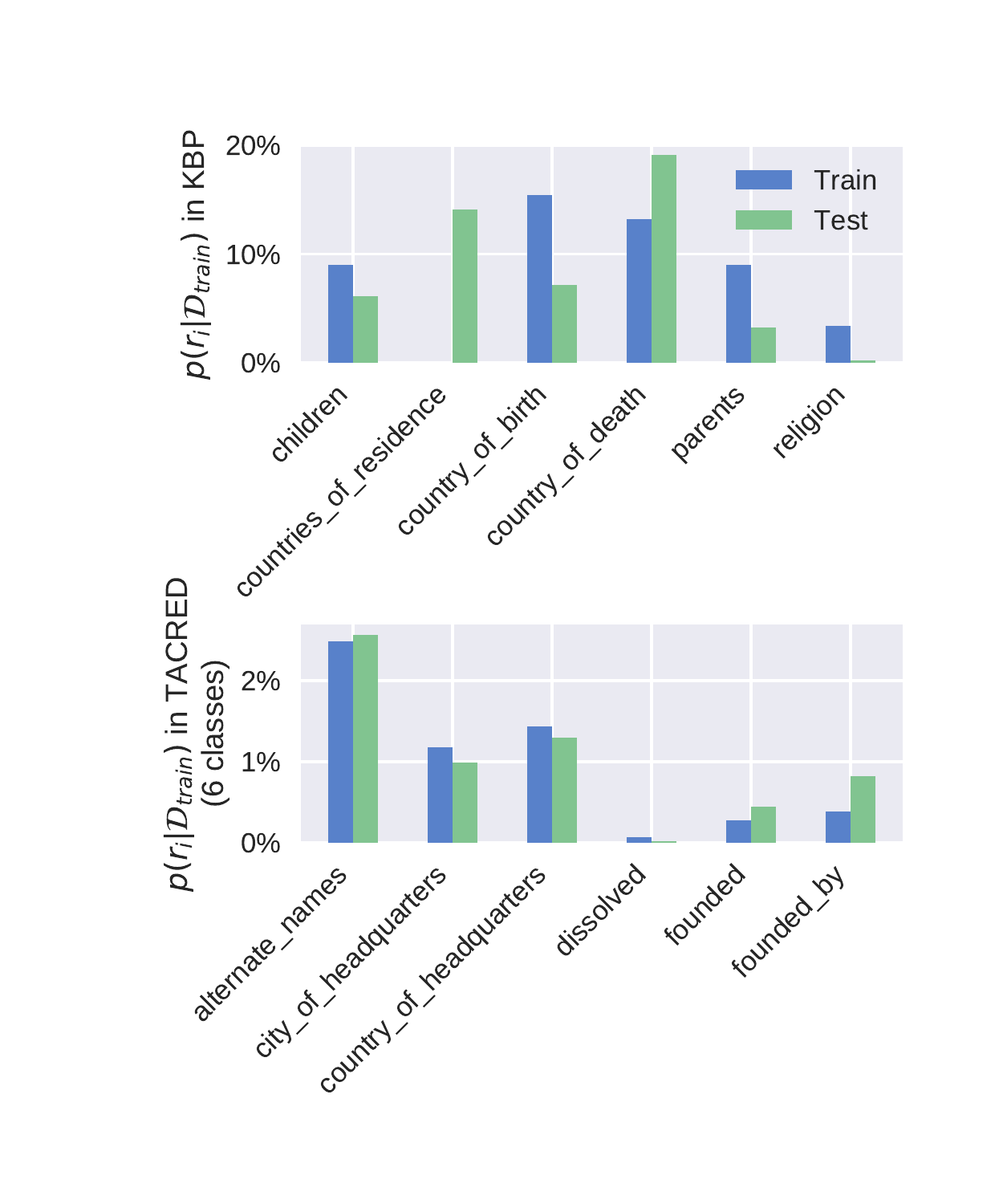}
\end{minipage}%
\begin{minipage}{.21\textwidth}
  \centering
  \includegraphics[trim={0.1cm, 1.5cm, 0cm, 2.0cm},clip,width=\linewidth]{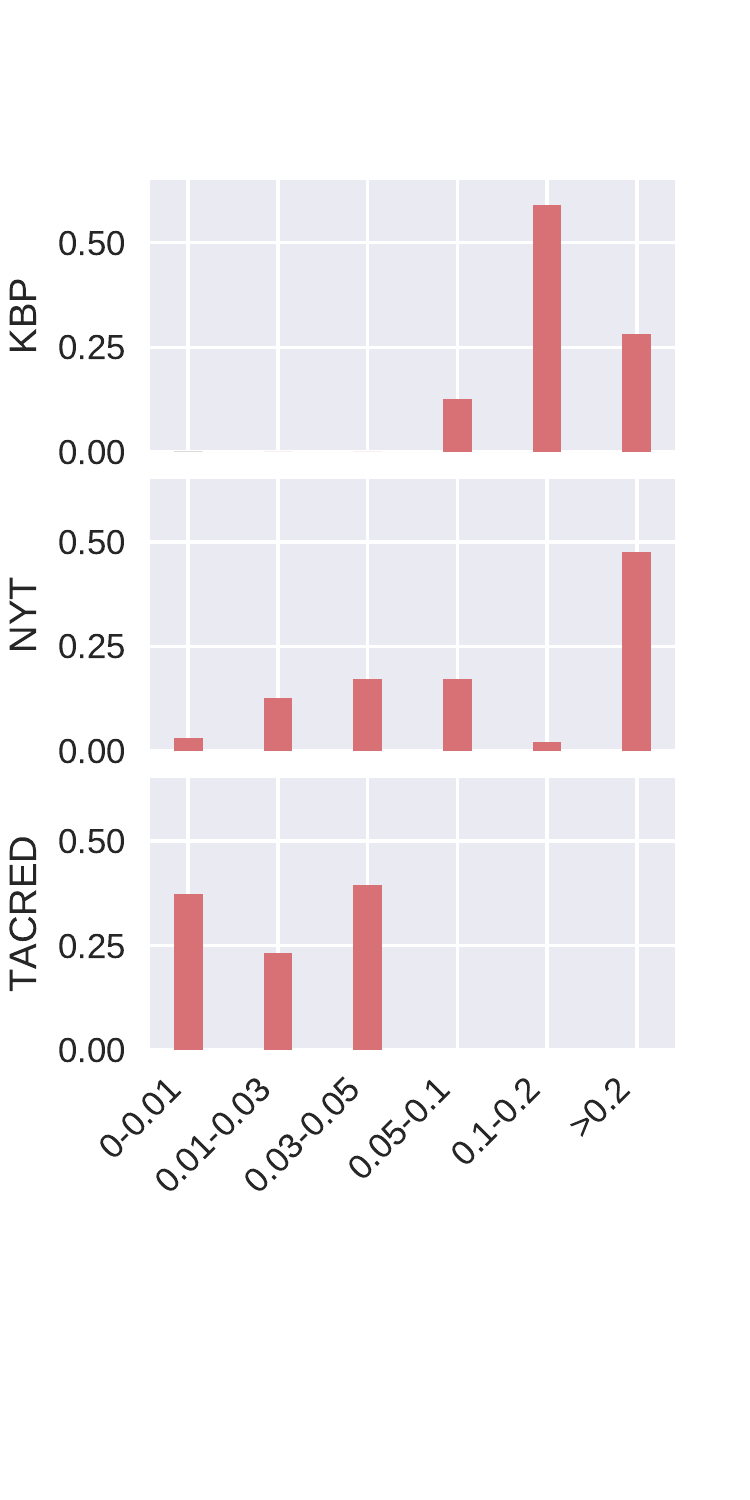}
\end{minipage}
\vspace{-0.2cm}
\caption{\textbf{Left:} Label distributions of KBP (distantly supervised dataset) are shifted, while those of TACRED (human-annotated dataset, 6 classes for illustration) are consistent. Full distributions for NYT and TACRED can be found in Appendix~\ref{sec:appendix}.
\textbf{Right:} Each relation $r_i$ is categorized into intervals along x-axis according to $\left|p(r_i|\mathcal{D}_{train})-p(r_i|\mathcal{D}_{test})\right|$; height of the bars indicates proportion of test instances that fall into each category, revealing that the shift is severe in DS datasets such as KBP and NYT.}
\label{fig:shifted_distrib}
\vspace{-0.5cm}
\end{figure}
Aiming to identify the relation among an entity pair, relation extraction (RE) serves as an important step towards text understanding and is a long-standing pursuit by many researchers.
To reduce the reliance on human-annotated data, especially for data-hungry neural models~\cite{zeng2014relation,zhang2017position}, there have been extensive studies on leveraging \textit{distant supervision} (DS) in conjunction with external knowledge bases to automatically generate large-scale training data~\cite{mintz2009distant,zeng2015distant}.
While recent DS-based relation extraction methods 
focus on handling \textit{label noise}~\cite{riedel2010modeling,hoffmann2011knowledge,lin2016neural}, \ie, false labels introduced by error-prone DS process,
other factors may have been overlooked. 
Here, we observe model behaviors to be different on DS datasets and clean dataset, which implies existence of other challenges that restrict performance of DS-RE models. 
In this paper, we conduct thorough analyses over both real-world and synthetic datasets to explore the question ---
\textit{what limits the performance of DS-trained neural models}.

Our analysis starts with a performance comparison among recent relation extraction methods on both DS datasets (\ie, KBP~\citep{ellis2012linguistic},  NYT~\citep{riedel2010modeling}) and human-annotated dataset (\ie, TACRED~\citep{zhang2017position}), with the goal of seeking models that can consistently yield strong results. 
We observe that, on human-annotated dataset, neural relation extraction models outperform feature-based models by notable gaps, but these gaps diminish when the same models are applied to DS datasets---neural models merely achieve performance comparable with feature-based models. 
We endeavor to analyze the underlying problem that leads to this unexpected ``diminishing'' phenomenon.

Inspired by two heuristic threshold techniques that prove to be effective on DS datasets, and further convinced by comprehensive analysis on synthetic datasets, we reveal an important characteristic of DS datasets---\textit{shifted label distribution}, the issue that the label distribution of training set does not align with that of test set.
There often exists a large margin between label distributions of \textit{distantly-supervised training set} and that of \textit{human-annotated test set}, as shown in Fig~\ref{fig:shifted_distrib}. 
Intuitively, this is mainly caused by ``false positive'' and ``false negative'' labels generated by error-prone DS processes, and the imbalanced data distribution of external knowledge bases. 

To some extent, such distortion is a special case of 
domain shift 
--- \ie, training the model on a source domain and applying the learned model to a different target domain. 
To further verify our assumption, we develop a simple domain adaption method, \textit{bias adaptation}, to address the shifted label distribution issue. 
It modifies the bias term in softmax classifiers and explicitly fits models along the shift. 
Specifically, the proposed method estimates the label distribution of target domain with a small development set sampled from test set, and derives the adapted predictions under reasonable assumptions.
In our experiments, we observe consistent performance improvement, which validates that model performance may be severely hindered by label distribution shift.

In the rest of the paper, we first introduce the problem setting in Section \ref{sec:exp} and report the inconsistency of model performance with human annotations and DS in Section \ref{sec:diminish}. 
Then, we present two threshold techniques which are found to be effective on DS datasets and further lead us to the discovery of shifted label distribution. 
We explore its impact on synthetic datasets in Section~\ref{simu}, and introduce the bias adjustment method in Section~\ref{adapt}. In addition, comparison of denoising method, heuristic threshold and bias adjustment is conducted in Section~\ref{comparison}. We discuss related work in Section~\ref{sec:related} and conclude our findings in Section~\ref{sec:conclusion}.

\section{Experiment Setup}\label{sec:exp}
In this paper, we conduct extensive empirical analyses on distantly supervised relation extraction (DS-RE). 
For a meaningful comparison, we ensure the same setup in all experiments.
In this section, we provide a brief introduction on the setting, while more details could be found in Appendix \ref{sec:appendix-exp}. 
All implementations are available at \url{https://github.com/INK-USC/shifted-label-distribution}.

\subsection{Problem Setting} 
Following previous works~\cite{ren2017cotype,liu2017heterogeneous}, we conduct relation extraction at \textit{sentence level}. 
Formally speaking, the basic unit is the relation mention, which is composed of one sentence and one ordered entity pair within the sentence. 
The relation extraction task is to categorize each relation mention into a given set of relation types, or a Not-Target-Type (\textsc{None}).

\subsection{Datasets}
We select three popular relation extraction datasets as benchmarks. Specifically, two of them are distantly supervised and one is human-annotated.

\noindent 
\textbf{KBP} \cite{ling2012fine} uses Wikipedia articles annotated with Freebase entries as train set, and manually-annotated sentences from 2013 KBP slot filling assessment results \cite{ellis2012linguistic} as test set.

\noindent 
\textbf{NYT} \cite{riedel2010modeling} contains New York Times news articles and has been already heuristically annotated. 
Test set is constructed manually by \cite{hoffmann2011knowledge}.

\noindent 
\textbf{TACRED} \cite{zhang2017position}  is a large-scale crowd-sourced dataset, and is sufficiently larger than previous manually annotated datasets. 

\begin{table}[t]
\footnotesize
\centering
\scalebox{0.98}{
\begin{tabularx}{1.05\columnwidth}{l*{3}{Y}}
\toprule
\multirow{2}{*}{Dataset} & \multicolumn{2}{c}{\textbf{\tiny Distantly Supervised}} & \multicolumn{1}{c}{\textbf{\tiny Human-annotated}} \\         & \multicolumn{1}{c}{KBP}  & \multicolumn{1}{c}{NYT}  & \multicolumn{1}{c}{TACRED} \\ \midrule
\#Relation Types  & 7        & 25     & 42          \\
\#Train Sentences & 23,784    & 235,982 & 37,311     \\
\#Test Sentences  & 289      & 395    & 6,277    \\ 
\bottomrule
\end{tabularx}
}
\caption{Statistics of Datasets Used in Our Study.}\label{tab-data}
\vspace{-0.2cm}
\end{table}

\begin{table*}[tb]
\centering
\scalebox{0.84}{%
\begin{tabularx}{1.1\linewidth}{ll|*{2}{Y}|*{1}{Y}}
\hline
& \multirow{2}{*}{\textbf{Method}~/~\textbf{Dataset}} & \multicolumn{2}{c|}{\textbf{Distantly-supervised}} & \multicolumn{1}{c}{\textbf{Human-annotated}} \\
& & \multicolumn{1}{c}{Wiki-KBP}  & \multicolumn{1}{c|}{NYT}  & \multicolumn{1}{c}{TACRED}  \\ \hline\hline
\multirow{3}{*}{Feature-based} & CoType-RM~\citep{ren2017cotype} & 28.98 $\pm$ 0.76 & 40.26 $\pm$ 0.51 & 45.97 $\pm$ 0.34 \\
 & ReHession~\cite{liu2017heterogeneous} & 36.07 $\pm$ 1.06 & 46.79 $\pm$ 0.75 & 58.06 $\pm$ 0.54 \\
& Logistic~\cite{mintz2009distant} & 37.58 $\pm$ 0.27 & 47.33 $\pm$ 0.44 & 51.67 $\pm$ 0.03\\
\hline
\multirow{7}{*}{Neural} & CNN~\cite{zeng2014relation} & 30.53 $\pm$ 2.26 & 46.75 $\pm$ 2.79 & 56.96 $\pm$ 0.43 \\
 & PCNN~\cite{zeng2015distant} & 31.58 $\pm$ 0.35 & 44.63 $\pm$ 2.70 & 58.39 $\pm$ 0.71 \\
 & Bi-GRU & \textbf{37.77 $\pm$ 0.18} & 47.88 $\pm$ 0.85 & 65.38 $\pm$ 0.60 \\
 & Bi-LSTM & 34.51 $\pm$ 0.99 & 48.15 $\pm$ 0.87 & 62.74 $\pm$ 0.23 \\
 & PA-LSTM~\cite{zhang2017position} & 37.28 $\pm$ 0.81 & 46.33 $\pm$ 0.64 & \textbf{65.69 $\pm$ 0.48} \\
& Bi-GRU-ATT~\cite{lin2016neural} & 37.50 $\pm$ 1.21 & \textbf{49.67  $\pm$ 1.06} & - \\
& PCNN-ATT~\cite{lin2016neural} & 33.74 $\pm$ 2.19 & 46.82 $\pm$ 0.82 & - \\

\hline
\end{tabularx}
}
\caption{\textbf{Performance Comparison of RE Models.} 5-time average and standard deviation of F1 scores are reported.}
\label{main-result}
\vspace{-0.4cm}
\end{table*}

\subsection{Pre-processing}
We leverage pre-trained GloVe \cite{pennington2014glove} embedding\footnote{\url{http://nlp.stanford.edu/data/glove.840B.300d.zip}}, and use the StanfordNLP toolkit~\cite{manning-EtAl:2014:P14-5} to get part of speech (POS) tags, named-entity recognition (NER) tags, and dependency parsing trees.

For the development sets, we use the provided development set on TACRED, and randomly split 10\% of training set on KBP and NYT. This development set is also refered to as \textit{noisy dev}, as we will introduce a \textit{clean dev} to deal with shifted label distribution in the later parts.

\subsection{Models}
We consider two 
classes of relation extraction methods, \ie, feature-based and neural models. 

Specifically, Feature-based models 
include
Logistic Regression~\citep{mintz2009distant}, CoType-RM~\citep{ren2017cotype} and  ReHession~\cite{liu2017heterogeneous}.
Neural models 
include
Bi-LSTMs and Bi-GRUs~\cite{zhang2017position}, Position-Aware LSTM~\cite{zhang2017position}, CNNs and PCNNs~\cite{zeng2014relation, zeng2015distant}.

For each relation mention, these models will first construct a representation vector $\h$, and then make predictions with softmax based on $\h$\footnote{CoType-RM and logistic regression are exceptions as they don't adopt softmax to generate output.}:
\begin{equation}
p(y = r_i|\h) = \frac{\exp(\r_i^T\h + b_i)}{\sum_{r_j}\exp(\r_j^T\h+b_j)},
\label{eqn:softmax}
\end{equation}
where $\r_i$ and $b_i$ are the parameters corresponding to \textit{i}-th relation type. 
More details on these models can be found in Appendix~\ref{sec:appendix-exp}.


\section{The Diminishing Phenomenon}\label{sec:diminish}

\begin{figure}
    \centering 
    \includegraphics[clip,width=0.42\textwidth]{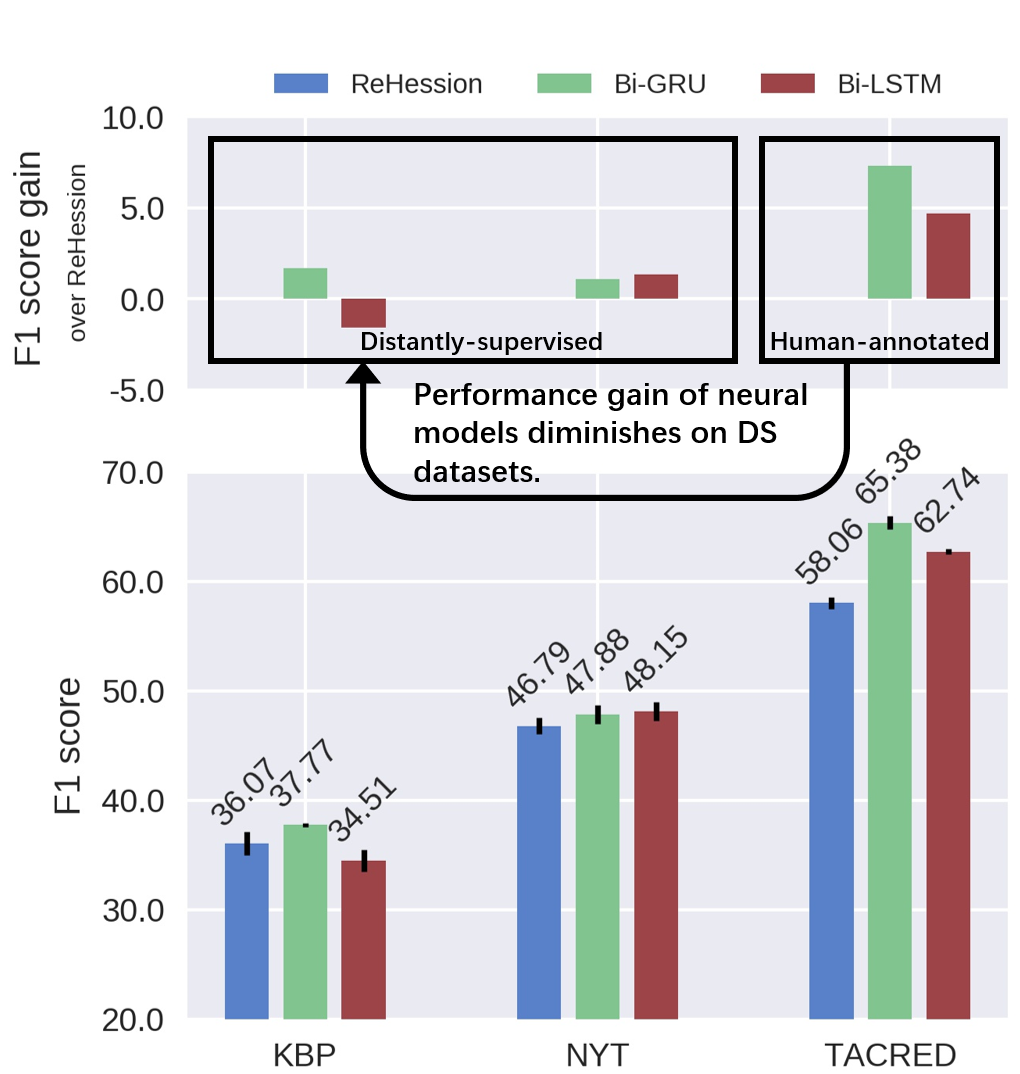}
    \vspace{-0.2cm}
    \caption{\textbf{F1 score comparison among three models.} Bi-GRU outperforms ReHession with a significant gap on TACRED, but only has comparable performance (with ReHession) on KBP and NYT. Similar gap diminishing phenomenon happens to Bi-LSTM.}
    \label{fig:diminishing}
    \vspace{-0.4cm}
\end{figure}

Neural models alleviate the reliance on hand-crafted features and 
have greatly advanced
the state-of-the-art, especially on datasets with human annotations. 
Meanwhile, we observe such performance boost starts to diminish on distantly supervised datasets. 
Specifically, we list the performance of all tested models in Table~\ref{main-result} and summarize three popular models in Figure~\ref{fig:diminishing}.  

On TACRED, a human annotated dataset, complex neural models like Bi-LSTM and Bi-GRU significantly outperform feature-based ReHession, with an up to 13\% relative F1 improvement. 
On the other hand, on distantly supervised datasets (KBP and NYT), the performance gaps between the aforementioned methods diminishes to within 5\% (relative improvement).
We refer to this observation as ``diminishing'' phenomenon. 
Such observation implies a lack of handling of the underlying difference between human annotations and distant supervision.

After a broad exploration, we found two heuristic techniques that we believe capture problems exclusive to distantly supervised RE, and are potentially related to ``diminishing'' phenomenon.
We found they can greatly boost the performance on distantly supervised datasets, but fail to do so on human-annotated dataset. 
To get deeper insights, we analyze the diminishing phenomenon and the two heuristic methods.


\begin{figure*}[t]
    \centering 
    \includegraphics[clip,width=\textwidth]{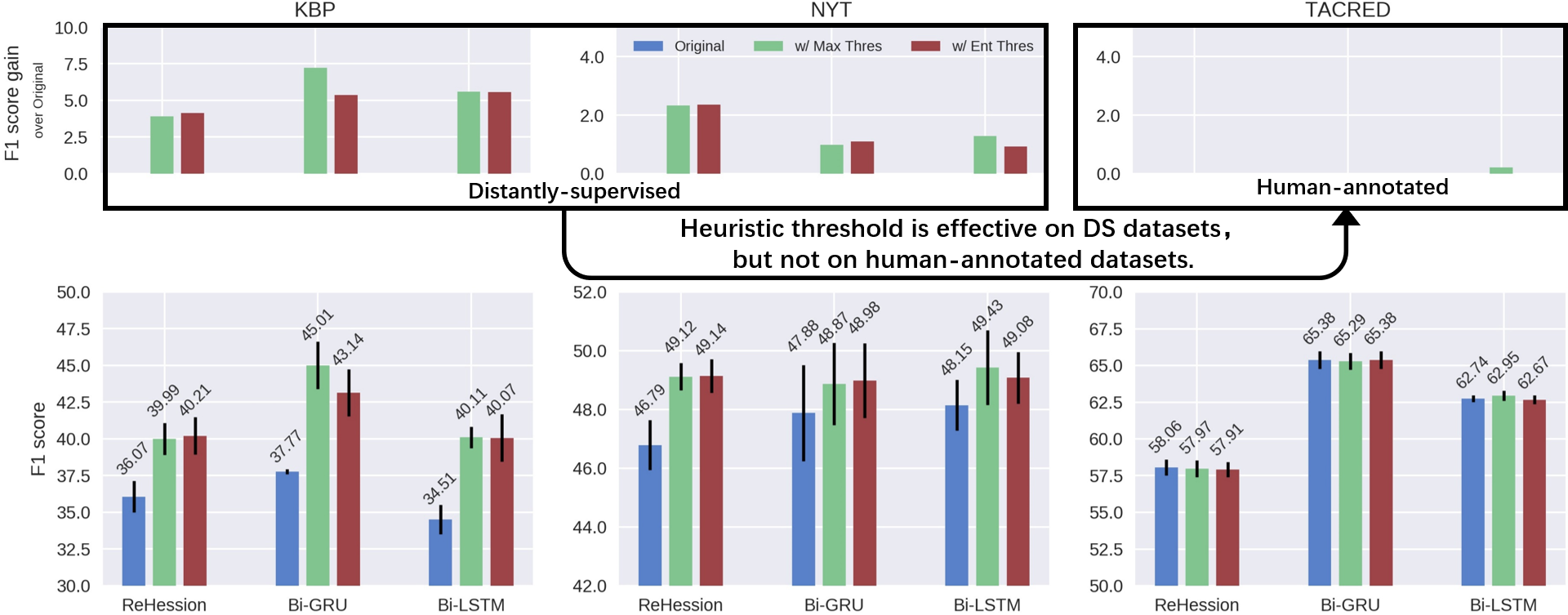}
    \caption{\textbf{F1 scores on three datasets when (a) no threshold (original) (b) max threshold (c) entropy threshold are applied, respectively.} A clear boost is observed on distantly supervised datasets (\ie KBP and NYT) after applying threshold; however, performance n human-annotated dataset (\ie TACRED) remains almost the same with thresholding. Heuristic threshold techniques may capture some important but overlooked problems in distantly supervised relation extraction.} 
    \label{fig:f1_with_thres}
    
\end{figure*}

\subsection{Heuristic Threshold Techniques}
Max threshold and entropy threshold are designed to identify the ``ambiguous'' relation mentions (\ie, predicted with a low confidence) and label them as the \textsc{None} type~\cite{ren2017cotype,liu2017heterogeneous}. In particular, referring to the original predictions as $r^* = \mathop{\arg\max}_{r_i} p(y=r_i|\mathbf{h})$, we formally introduce these two threshold techniques:
\begin{itemize}[leftmargin=10pt]
    \item \textbf{Max Threshold} introduces an additional hyper-parameter $T_m$, and adjusts the prediction as~\cite{ren2017cotype}:
    \begin{equation*}
    predict(\mathbf{h})=\left\{
    \begin{aligned}
    &r^*, && p(y=r^*|\mathbf{h})>T_m \\
    &\textsc{None}, && \mbox{Otherwise}
    \end{aligned}
    \right..
    \end{equation*}
    
    \item \textbf{Entropy Threshold} introduces an additional hyper-parameter $T_e$. It first calculates the entropy of prediction:
    \begin{equation*}
        e(\mathbf{h})= - \sum_{r_k} p(y=r_k|\mathbf{h}) \log p(y=r_k|\mathbf{h}),
    \end{equation*}
    then it adjusts prediction as~\cite{liu2017heterogeneous}:
    \begin{equation*}
    predict(\mathbf{h})=\left\{
    \begin{aligned}
    &r^*, && e(\mathbf{h})<T_e \\
    &\textsc{None}, && \mbox{Otherwise}
    \end{aligned}
    \right..
    \end{equation*}
\end{itemize}

To estimate $T_e$ or $T_m$, 20\% instances are sampled from the test set as an additional development set, and used to tune the value of $T_e$ and $T_m$ with grid search.
After that, we evaluate the model performance on the remaining 80\% of the test set. 
We refer to this new dev set as \textit{clean dev} and refer to the original dev set used for tuning other parameters as \textit{noisy dev}. We would like to highlight that tuning threshold on this clean dev is necessary as it acts as a bridge between distantly supervised train set and human-annotated test set.

\subsection{Results and Discussion}

Results of three representative models (ReHession, Bi-GRU and Bi-LSTM) using threshold techniques are summarized in Figure~\ref{fig:f1_with_thres}. Full results are listed in Table~\ref{table-bias}.
We observe significant improvements on distantly supervised datasets (\ie, KBP and NYT), with a up to $19\%$ relative F1 improvement (Bi-GRU from 37.77\% to 45.01\% on KBP). 
However, on the human-annotated corpus, the performance gain can be hardly noticed. 
Such inconsistency implies that these heuristics may capture some important but overlooked factors for distantly supervised relation extraction, while we are still unclear about their underlying mechanisms. 

\label{false}
Intuitively, annotations provided by distant supervision differ from human annotations in two ways: (1) \textbf{False Positive:} falsely annotating unrelated entities in one sentence as a certain relation type; (2) \textbf{False Negative:} neglecting related entities by marking their relationship as \textsc{None}. 
The terms, false positive and false negative, are commonly used to describe wrongful predictions. Here we only borrow the terms for describing label noises.
These label noises distort the true label distribution of train corpora, creating a gap between the label distribution of train and test set (\ie, shifted label distribution).
With existing denoising methods, the effect of noisy training instances may be reduced;
still, it would be infeasible to recover the original label, and thus label distribution shift remains an unsolved problem.

In our experiments, we notice that in distantly supervised datasets, instances labeled as \textsc{None} have a larger portion in test set than in train set. It is apparent that the strategy of rejecting ``ambiguous'' predictions would guide the model to predict more \textsc{None} types, leading the predicted label distribution towards a favorable direction.
Specifically, in the train set of KBP, 74.25\% instances are annotated as \textsc{None} and 85.67\% in the test set. 
The original Bi-GRU model would annotate 75.72\% instances to be \textsc{None}, which is close to 74.25\%; after applying the max-threshold and entropy-threshold, this proportion becomes 86.18\% and 88.30\%, which are close to 85.67\%.  

Accordingly, we believe part of the underlying mechanism of heuristic threshold is to better handle the label distribution difference between train and test set. We try to further verify this hypothesis with experiments in next section.

\section{Shifted Label Distribution}
\label{simu}

In this section, we first summarize our observation on shifted label distribution, and then conduct empirical analysis to study its impact on model performance using synthetic datasets. These datasets are carefully created so that label distribution is the only variable and other factors are controlled.  

\subsection{Shifted Label Distribution}

Shifted label distribution refers to the problem that the label distribution of train set does not align with the test set. 
This problem is related to but different from ``learning from imbalanced data'', 
where the data distribution is imbalanced, but consistent across train and test set.
Admittedly, one relation may appear more or less than another in natural language, creating distribution skews; however, this problem widely occurs in both supervised and distantly supervised settings, and is not our focus in this paper. 

Our focus is the \textit{label distribution difference between train and test set}.
This problem is critical to distantly supervised relation extraction, where the train set is annotated with distant supervision and the test set is manually annotated. 
As previously mentioned in Section~\ref{false}, distant supervision differs from human annotations by introducing ``false positive'' and ``false negative'' labels. The label distribution of train set is subject to existing entries in KBs, and thus there exists a gap between label distributions of train and test set.

We visualize the distribution of KBP (DS dataset) and a truncated 6-class version of TACRED (human-annotated dataset) in Fig~\ref{fig:shifted_distrib}. Also, proportions of instances categorized into $\delta = \left|p(r_i|\mathcal{D}_{train}) - p(r_i|\mathcal{D}_{test})\right|$ bins are shown.
It is observed that KBP and NYT both have shifted label distributions (most instances fall into $\delta >0.05$ bins); while TACRED has consistent label distributions (all instances fall into $\delta < 0.05$ bins). 

\subsection{Impact of Shifted Label Distribution}

In order to quantitatively study the impact of label distribution shift, we construct synthetic datasets by sub-sampling instances from the human-annotated TACRED dataset. 
In this way, the only variable is the label distribution of synthetic datasets, and the impact of other factors such as label noise is excluded.

\smallsection{Synthetic Dataset Generation}
We create five synthetic train sets, by sampling sentence-label pairs from original TACRED train set with label distributions $\mathsf{S1}$-$\mathsf{S5}$. 
$\mathsf{S5}$ is a randomly generated label distribution (see Fig~\ref{fig:TACRED original} in Appendix~\ref{sec:appendix}). 
$\mathsf{S0}$ is TACRED's original train set label distribution.
$\mathsf{S1-S4}$ are created with linear interpolation between $\mathsf{S0}$ and $\mathsf{S5}$, \textit{\ie}, $\mathsf{Si} = \frac{5-i}{5}\mathsf{S0} + \frac{i}{5}\mathsf{S5}$. 
We apply disproportionate stratification to control the label distribution of synthetic datasets.
In this way, we ensure that the number of sampled instances in each synthetic train set is kept constant within $10000\pm 3$, and the label distribution of each set satisfies $\mathsf{S1}$-$\mathsf{S5}$ respectively. 

\begin{figure}[t]
    \vspace{-0.5cm}
    \centering
    \includegraphics[trim={1.8cm 1cm 1.8cm 1cm},clip,width=0.48\textwidth]{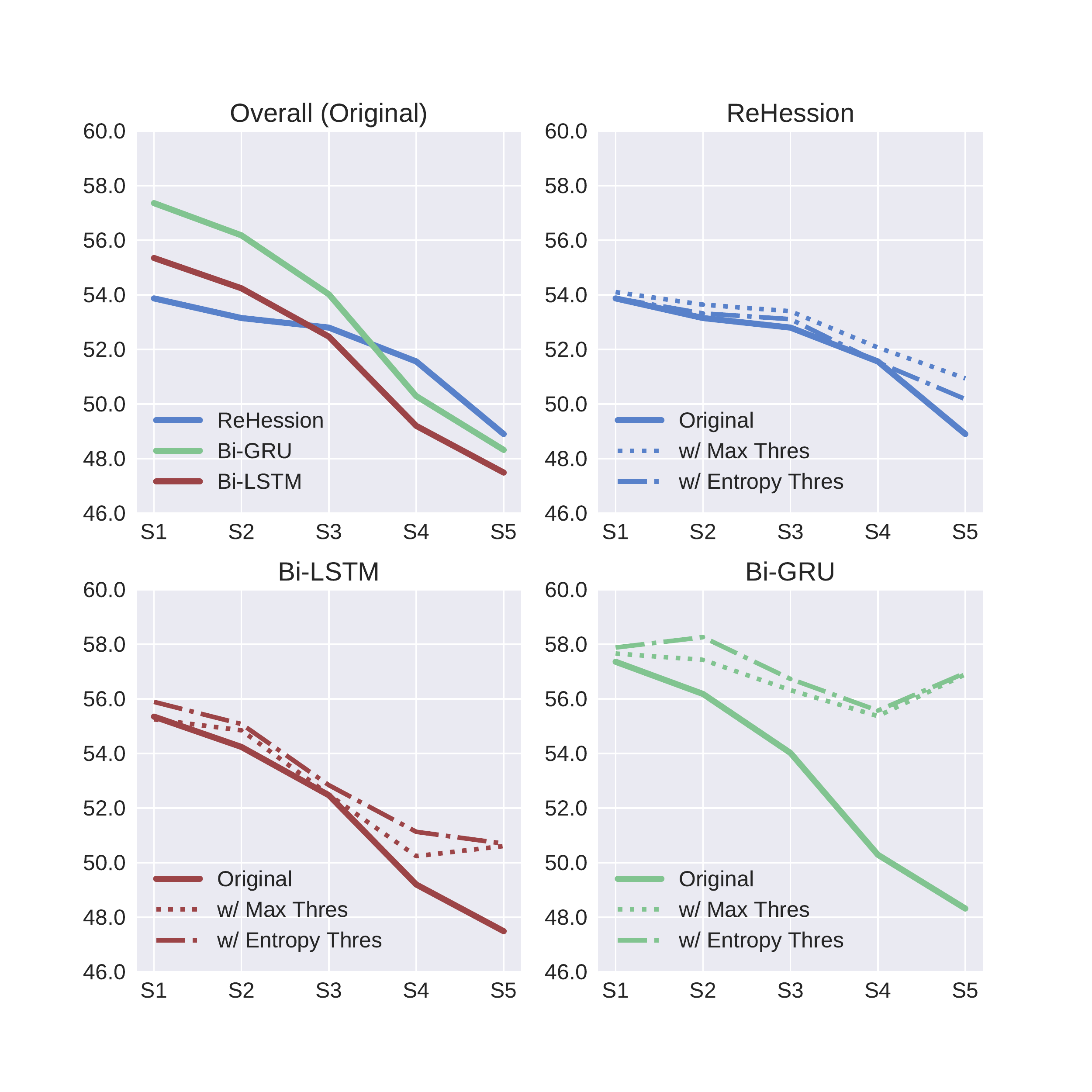}
    \vspace{-0.8cm}
    \caption{\textbf{F1 scores on synthesized datasets $\mathsf{S1}$-$\mathsf{S5}$.} We observe that (a) performance consistently drops from $\mathsf{S1}$ to $\mathsf{S5}$, demonstrating the impact of shifted label distributions; (b) ReHession is more robust to such distribution shift, outperforming Bi-LSTM and Bi-GRU on $\mathsf{S4}$ and $\mathsf{S5}$; (c) threshold is an effective way to handle such shift.}
    \label{fig:imbalanced-f1}
    \vspace{-0.2cm}
\end{figure}

\smallsection{Results and Discussion}
We conduct experiments with three typical models (\ie, Bi-GRU, Bi-LSTM and ReHession) and summarize their results in Fig~\ref{fig:imbalanced-f1}. 
We observe that, from $\mathsf{S1}$ to $\mathsf{S5}$, the performance of all models consistently drops.
This phenomenon verifies that shifted label distribution is making a negative influence on model performance. The negative effect expands as the train set label distribution becomes more twisted.

At the same time, we observe that feature-based ReHession is more robust to such shift.
The gap between ReHession and Bi-GRU stably decreases, and eventually ReHession starts to outperform the other two at $\mathsf{S4}$. 
This could be the reason accounting for ``diminishing'' phenomenon --- neural models such as Bi-GRU is supposed to outperform ReHession by a huge gap (as with $\mathsf{S1}$); however on distantly supervised datasets, shifted label distribution seriously interfere the performance (as with $\mathsf{S4}$ and $\mathsf{S5}$), and therefore it appears as the performance gap diminishes.

\smallsection{Applying Threshold Techniques}
We also applied the two threshold techniques on the synthetic datasets and summarize their performance in Fig~\ref{fig:imbalanced-f1}. 
The three models become more robust to the label distribution shift when threshold is applied. 
Threshold techniques are consistently making improvements, and the improvements are more significant with $\mathsf{S5}$ (most shifted) than with $\mathsf{S0}$ (least shifted).
This observation verifies that the underlying mechanism of threshold techniques help the model better handle label distribution shift.

\section{Bias Adjustment: An Adaptation Method for Label Distribution Shift}
\label{adapt}

\begin{table*}[htbp!]
\centering 
\scalebox{0.75}{
\begin{tabular}{llccccccccc}
\hline
Dataset  & Model  & Original       & Max-thres & $\Delta$  & Ent-thres & $\Delta$ & BA-Set & $\Delta$ & BA-Fix  & $\Delta$   \\ \hline\hline
\multirow{6}{*}{KBP} & ReHession & 36.07 $\pm$ 1.06 & 39.99 $\pm$ 1.09 & 3.92 & 40.21 $\pm$ 1.28 & 4.14 & 37.18 $\pm$ 1.59 & 1.11 & 37.86 $\pm$ 1.64 & 1.79\\
    & Bi-GRU       & 37.77 $\pm$ 0.18 & 45.01 $\pm$ 1.61 & 7.24 & 43.14 $\pm$ 1.59 & 5.37 & 40.63 $\pm$ 1.79 & 2.86 & 42.28 $\pm$ 2.05 & 4.51 \\
    & Bi-LSTM      & 34.51 $\pm$ 0.99 & 40.11 $\pm$ 0.73 & 5.60 & 40.07 $\pm$ 1.61 & 5.56 & 37.43 $\pm$ 1.72  & 2.92 & 38.28 $\pm$ 1.66 & 3.77\\
    & PA-LSTM   & 37.28 $\pm$ 0.81 & 44.04 $\pm$ 1.09 & 6.76 & 43.17 $\pm$ 1.25 & 5.89 & 40.47 $\pm$ 2.55 & 3.19 & 42.44 $\pm$ 1.32 & 5.16\\
    & CNN       & 30.53 $\pm$ 2.26 & 31.23 $\pm$ 1.68 & 0.70 & 31.39 $\pm$ 2.43 & 0.86 & 30.73 $\pm$ 3.96 & 0.20 & 36.12 $\pm$ 2.65 & 5.59\\
    & PCNN      & 31.58 $\pm$ 0.35 & 32.98 $\pm$ 0.89 & 1.40 & 32.19 $\pm$ 0.76 & 0.61 & 32.79 $\pm$ 1.95 & 1.21 & 38.78 $\pm$ 2.63 & 7.20\\
    \hline
\multirow{6}{*}{NYT} & ReHession & 46.79 $\pm$ 0.75 & 49.12 $\pm$ 0.47  & 2.33 & 49.14 $\pm$ 0.57 & 2.35 & 48.50 $\pm$ 1.23 & 1.71 & 48.14 $\pm$ 1.08 & 1.35 \\
    & Bi-GRU       & 47.88 $\pm$ 0.85 & 48.87 $\pm$ 1.40  & 0.99 & 48.98 $\pm$ 1.27 & 1.10 & 48.40 $\pm$ 1.52 & 0.52 & 48.74 $\pm$ 1.94 & 0.86 \\
    & Bi-LSTM      & 48.15 $\pm$ 0.87 & 49.43 $\pm$ 1.27 & 1.28 & 49.08 $\pm$ 0.88 & 0.93 & 49.72 $\pm$ 1.24 & 1.57 & 49.90 $\pm$ 1.44 & 1.75\\
    & PA-LSTM   & 46.33 $\pm$ 0.64 & 46.70 $\pm$ 1.43 & 0.37 & 48.65 $\pm$ 1.24 & 2.32  & 47.77 $\pm$ 1.43 & 1.44 & 48.65 $\pm$ 1.24 & 2.32\\
    & CNN       & 46.75 $\pm$ 2.79 & 47.87 $\pm$ 1.89 & 1.12 & 46.54 $\pm$ 2.70 & -0.21 & 48.83 $\pm$ 1.66 & 2.08 & 48.17 $\pm$ 1.88 & 1.42\\
    & PCNN      & 44.63 $\pm$ 2.70 & 48.31 $\pm$ 0.40 & 3.68 & 47.04 $\pm$ 2.60 & 2.41 & 49.08 $\pm$ 0.88 & 4.45 & 48.72 $\pm$ 1.72 & 4.09\\ 
    \hline
\multirow{3}{*}{TACRED} & ReHession & 58.06 $\pm$ 0.54 & 57.97 $\pm$ 0.57 & -0.09 & 57.91 $\pm$ 0.52 & -0.15 & 58.59 $\pm$ 0.66 & 0.53 & 58.61 $\pm$ 0.99 & 0.55\\
    & Bi-GRU       & 65.38 $\pm$ 0.60 & 65.29 $\pm$ 0.58 & 0.01 & 65.38 $\pm$ 0.60 & 0.10 &63.72 $\pm$ 0.64 & -1.56 & 64.70 $\pm$ 0.46 & -0.68\\
    & Bi-LSTM      & 62.74 $\pm$ 0.23 & 62.95 $\pm$ 0.35 & 0.21 & 62.67 $\pm$ 0.29 & -0.07 & 62.73 $\pm$ 0.60 & -0.01 & 63.44 $\pm$ 0.54 & 0.70\\
	\hline
\end{tabular}
}
\caption{\textbf{F1 score of RE Models with Threshold and Bias Adaptation.} 5-time average and standard deviation of F1 scores are reported. $\Delta$ denotes the F1 improvement over original. On DS datasets, the four methods targeting label distribution shift achieve consistent performance improvement, with averagely 3.83 F1 improvement on KBP and 1.72 on NYT. However, the same four methods fail to improve performance on human-annotated TACRED.} 
\label{table-bias}
\vspace{-0.2cm}
\end{table*}

Investigating the probabilistic nature of softmax classifier, we present a principled domain adaptation approach, \textit{bias adjustment} (BA), to deal with label distribution shift. This approach explicitly fits the model along such shift by adjusting the bias term in softmax classifier.

\subsection{Bias Adjustment}

We view corpora with different label distributions as different domains. 
Denoting distantly supervised corpus (train set) as $\D_d$ and human-annotated corpus (test set) as $\D_m$, our task becomes to calculate $p(y = r_i|\h,\D_m)$ based on $p(y = r_i|\h,\D_d)$. 

We assume the only difference between $p(y = r_i|\h,\D_m)$ and $p(y= r_i|\h,\D_d)$ to be the label distribution, and the semantic meaning of each label is unchanged. 
Accordingly, we assume $p(\h|r_i)$ is universal, \ie, 
\begin{equation}
    p(\h|r_i, \D_m) = p(\h|r_i, \D_d) = p(\h|r_i).
    \label{eqn:universal}
\end{equation}

As distantly supervised relation extraction models are trained with $\D_d$, its prediction in Equation~\ref{eqn:softmax} can be viewed as $p(y=r_i|\h,\D_d)$, \ie, 
\begin{align}
    p(y=r_i|\h,\D_d) &= p(y=r_i|\h) \nonumber\\  
    &= \frac{\exp(\r_i\Trans\h+b_i)}{\sum_{r_j} \exp(\r_j\Trans\h+b_j)}.\label{eq:bias0}
\end{align}

Based on the Bayes Theorem, we have:
\begin{equation}\small
p(y=r_i|\h,\D_m)=\frac{p(\h|r_i, \D_m) p(r_i|\D_m)}{\sum_{r_j}p(\h|r_j, \D_m)p(r_j|\D_m)}.\label{eq:bias1}
\end{equation}

Based on the definition of conditional probability and Equation~\ref{eqn:universal}, we have:
\begin{align}
p(\h|r_i, \D_m) &= p(\h|r_i)= p(\h|r_i, \D_d) \nonumber\\
&=\frac{p(\h, r_i, \D_d)}{p(r_i, \D_d)}\nonumber\\
&=\frac{p(r_i|\h, \D_d)\cdot p(\h |\D_d)}{p(r_i|\D_d)}.
\label{eq:bias2}    
\end{align}

With Equation~\ref{eq:bias0}, \ref{eq:bias1} and \ref{eq:bias2}, we can derive that
\begin{align}
&p(y=r_i|\h,\D_m)\nonumber\\
=&\frac{p(y=r_i|\h,\D_d)\dfrac{p(r_i|\D_m)}{p(r_i|\D_d)}}{\sum_{r_j} p(y=r_j|\h,\D_d)\dfrac{p(r_j|\D_m)}{p(r_j|\D_d)}}\nonumber\\
=&\frac{\exp(\r_i\Trans\h+b'_i)}{\sum_j\exp(\r_j\Trans\h+b'_j)},\label{eq:bias3}
\end{align}
where
\begin{equation}
    b'_i = b_i + \ln p(r_i|\D_m) - \ln p(r_i|\D_d)\label{eq:set_bias}.
\end{equation}

With this derivation we now know that, under certain assumptions (Equation~\ref{eqn:universal}), we can adjust the prediction to fit a target label distribution given $p(r_i|\D_d)$ and $p(r_i|\D_m)$. 
Accordingly, we use Equation~\ref{eq:bias3} and \ref{eq:set_bias} to calculate the adjusted prediction as:
\begin{equation}
r^* = \mathop{\arg\max}_{r_i} \exp(\r_i^T\h+b'_i).
\end{equation}

\noindent
\textbf{Label distribution estimation.}
The source domain (train) label distribution, $p(r_i|\D_d)$ can be easily estimated on train set.
As for target domain (test) distribution $p(r_i|\D_m)$, we use maximum likelihood estimation on a held-out \textit{clean dev} set, which is a 20\% sample from test set. 
This setting is similar to heuristic threshold techniques.

\noindent
\textbf{Implementation details.}
The bias adjustment model are implemented in two ways:
\begin{itemize}[leftmargin=10pt]
\item \textbf{BA-Set} directly replaces the bias term in Equation~\ref{eqn:softmax} with $b'_i$ in Equation~\ref{eq:set_bias} during evaluation.
No modification to model training is required.

\item \textbf{BA-Fix} fixes the bias term in Equation~\ref{eqn:softmax} as $b_i = \ln p(r_i|\D_d)$ during training and replaces it with $b'_i = \ln p(r_i|\D_m)$ during evaluation.
Intuitively, BA-Fix would encourage the model to fit our assumption better (Equation~\ref{eqn:universal}); still, it needs special handling during model training, which is a minor disadvantage of BA-Fix compared with BA-Set.
\end{itemize}

\subsection{Results and Discussion}
\begin{figure}[t]
    \centering
    \includegraphics[trim={0.3cm 0cm 1.8cm 1.0cm},clip,width=0.48\textwidth]{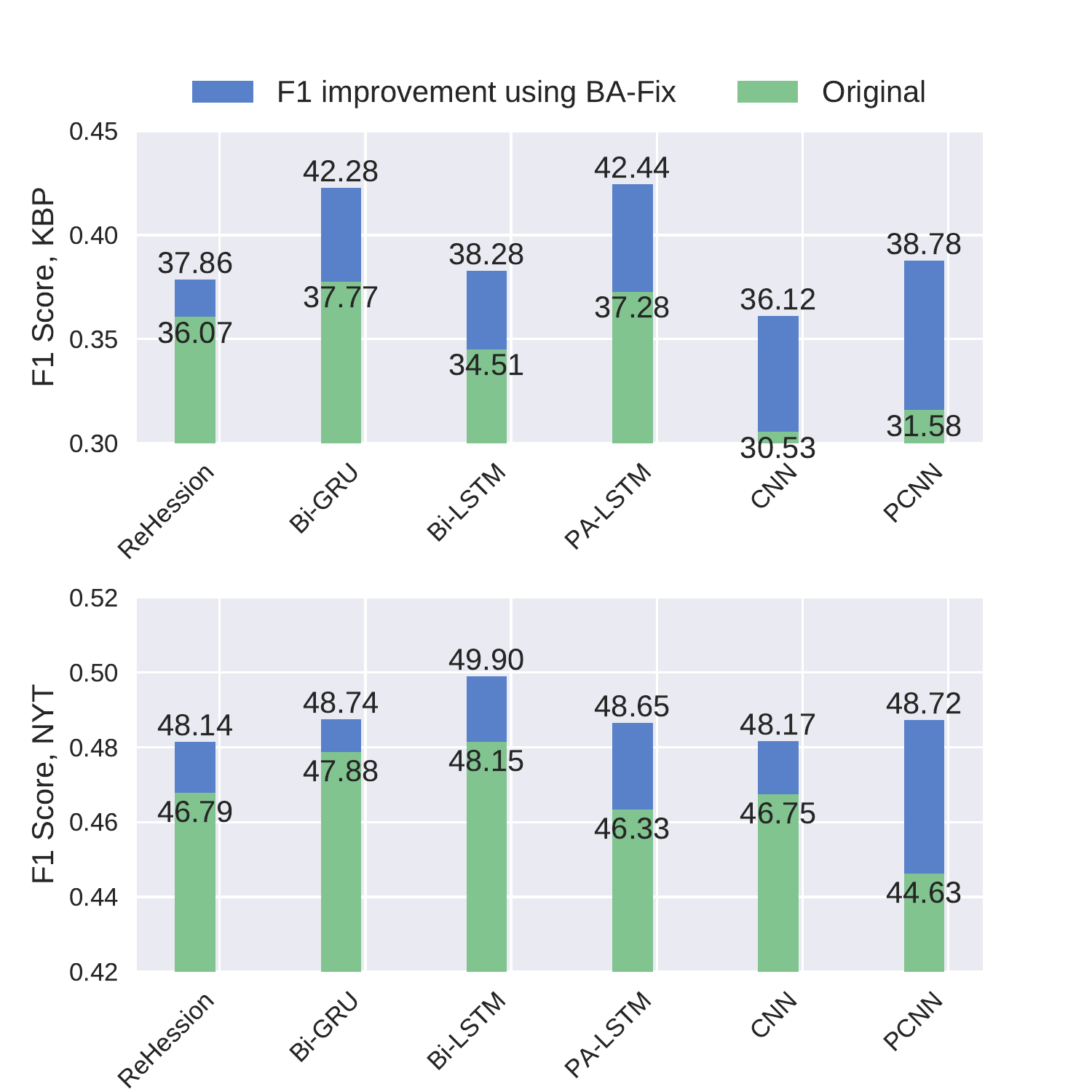}
    \vspace{-0.4cm}
    \caption{\textbf{F1 Improvement using BA-Fix.} BA-Fix consistently improves performance in compared models.}\label{fig:fig5}
    \vspace{-0.4cm}
\end{figure}

We conduct experiments to explore the effectiveness of BA-Set and BA-Fix and summarize their performance in Table \ref{table-bias}. Also, F1 improvements when BA-Fix is used is shown in Fig \ref{fig:fig5}.
We find that these two technologies bring consistent improvements to all RE models on both distantly supervised datasets. 
Especially, in the case of PCNN on KBP dataset, a $23\%$ relative F1 improvement is observed. 
At the same time, the same technology fails to achieve performance improvements on TACRED, the human annotated dataset. 
In other words, we found that, by explicitly adapting the model along label distribution shift, consistent improvements can be achieved on distant supervision but not on human annotations. 
This observation again supports our assumption that shifted label distribution is an unique factor for distantly supervised RE that needs special handling. 

\smallsection{Summary of all four methods}
Compared with heuristic threshold techniques, bias adjustment methods directly address shifted label distribution issue by adjusting bias term in classifier, and is more principled and explainable. Though both threshold and bias adjustment methods require an extra clean dev sampled from test set, threshold techniques require predicting on the instances in the clean dev and tune $T_m$ or $T_e$ based on the output probability. Bias adjustment only needs the label distribution estimated on clean dev, and is more efficient in computation.
As for performance, there is no clear indication that one method is consistently stronger than another.
However, all four methods (2 threshold and 2 bias adjustment) achieve similar results when applied to RE models -- we observe relatively significant performance improvements on DS datasets, but only marginal improvements on human-annotated dataset.

\smallsection{Comments on neural RE models}
On average, bias adjustment methods results in 3.66 F1 improvement on KBP and 2.05 on NYT for neural models. BA-Fix gains a suprising 7.20 with PCNN on KBP.
Noting that only bias terms in softmax classifier are modified and only a small piece of extra information is used, 
it is implied
that shifted label distribution is severely hindering model performance and capabilities of state-of-the-art neural models are not fully described with traditional evaluation.
Hidden representations $\h$ learned by neural models indeed capture semantic meanings more accurately than feature-based model, while the bias in classifier becomes the major obstacle towards better performance.


\section{Comparison with Denoising Methods}\label{comparison}

In this section, we conduct analyses about shifted label distribution and its relation with label noise. 
We apply a popular label noise reduction method--selective attention~\cite{lin2016neural}, which groups all sentences with the same entity pair into one bag, conducts multi-instance training and tries to place more weight on high-quality sentences within the bag. 
This method, 
along with threshold techniques and bias adjustment introduced in previous sections, 
is
applied 
to two different models (\ie, PCNN and Bi-GRU). 

\begin{figure}
    \centering
    \includegraphics[trim={0.2cm 0.5cm 0cm 0.5cm},clip,width=0.5\textwidth]{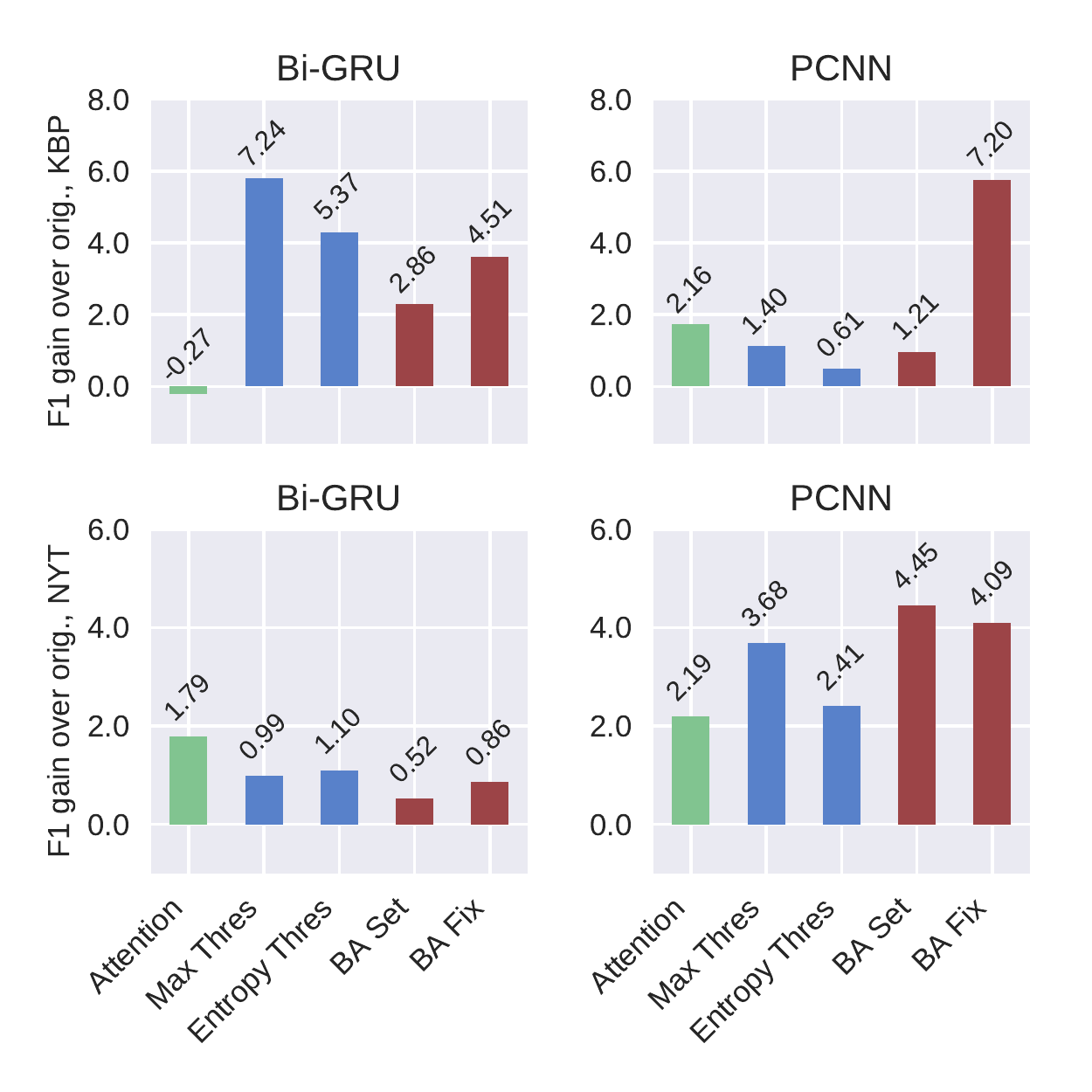}
    \vspace{-0.4cm}
    \caption{\textbf{Comparison among selective attention, threshold heuristics and bias adaption approaches.} Threshold heuristics and bias adaption approaches bring more significant improvements in some cases, indicating that shifted label distribution is a non-negligible problem.}
    \label{fig:corpus_bar}
    \vspace{-0.1cm}
\end{figure}

We summarize their improvements over the original model in Figure~\ref{fig:corpus_bar}. 
We find that selective attention is indeed effective and improves the performance; meanwhile, heuristic threshold and bias adaption approaches boost the performance, and in some cases the boost is even more significant than that of selective attention.
This observation is reasonable since both heuristics and bias adaption approaches are able to access additional information from clean dev (20\% of test set). 
Still, it is surprising that such small piece of information brings about huge difference, demonstrating the importance of handling shifted label distribution. 
It also shows that there exists much space for improving distantly supervised RE models from a shifted label distribution perspective.

\section{Related Work}\label{sec:related}

\subsection{Relation Extraction}
Relation extraction is to identify the relationship between a pair of entity mentions. The task setting slightly varies as the relation can be either extracted from a bag of sentences (corpus-level) or one single sentence (sentence-level).
In this paper, we focus on sentence-level RE. That is, prediction should be purely based on the information provided within the sentence, instead of external knowledge or commonsense.

Recent approaches follow the supervised learning paradigm and rely on considerable amounts of labeled instances to train effective models.
\citet{zeng2014relation} proposed using CNN for relation extraction, which could automatically capture features from texts. \citet{zeng2015distant} further extended it with piecewise max-pooling, \ie, splitting the sentence into three pieces with the object and subject entity, doing max-pooling over the three pieces separately, and finally concatenating the hidden representations. \citet{lin2016neural} applied selective attention over sentences for learning from multiple instances. This method was originally designed for corpus-level setting. It organizes sentences into bags and assign lower weights to those less relevant sentences in the bag. \citet{zhang2017position} proposed position-aware LSTM network which incorporates entity position information into encoding and enables attention mechanism to simultaneously exploit semantic and positional information.

\subsection{Distant Supervision}
In supervised relation extraction paradigm, one longstanding bottleneck is the 
lack of large-scale labeled training data. 
In order to alleviate the dependency on human supervision, \citet{mintz2009distant} proposed distant supervision, namely constructing large datasets automatically by aligning text to an existing knowledge bases (\eg, Freebase). Also, distantly supervised relation extraction is formulated into a reinforcement learning problem by \citet{feng2018reinforcement} for selecting high-quality instances. 
Similar annotation generation strategy using distant supervision has also been used for other NLP tasks, such as named entity recognition~\cite{shang2018learning} and sentiment classification~\cite{go2009twitter}.

Though this strategy lightens annotation burdens, distant supervision inevitably introduces label noises. 
As the relation types are annotated merely according to entity mentions in the sentence, the local context may be annotated with labels that are not expressed in the sentence, 
In recent years, researchers mainly focus on dealing with label noises, and proposed the following methods: \citet{riedel2010modeling} use multi-instance single-label learning paradigm;  \citet{hoffmann2011knowledge, surdeanu2012multi} propose multi-instance multi-label learning paradigm. 
Recently, with the advance of neural network techniques, deep learning methods \cite{zeng2015distant, lin2016neural} are applied to distantly supervised datasets, with powerful automatic feature extraction and advanced label noised reducing techniques such as selective attention. 
\citet{liu2017heterogeneous} proposed a general framework to consolidate heterogeneous information and refine the true labels from noisy labels. 

Label noise is certainly an important factor limiting the performance of DS-RE models. Meanwhile, we argue that shifted label distribution is also a performance-limiting aspect. It is long-overlooked and should be handled properly.


\section{Conclusion and Future Work}

In this paper, we first present the observation of inconsistent performance when models are trained with human annotations and distant supervision in the task of relation extraction.
It leads us to explore the underlying challenges for distantly supervised relation extraction. 
Relating two effective threshold techniques to label distribution, we reveal an important yet long-overlooked issue -- \textit{shifted label distribution}. 
The impact of this issue is further demonstrated with experiments on five synthetic train sets.
We also consider this issue from a domain adaptation perspective, introducing a theoretically-sound bias adjustment method to recognize and highlight label distribution shift. 
The bias adjustment methods achieve significant performance improvement on distantly-supervised datasets. 
All of these findings support our argument that shifted label distribution can severely hinder model performance and should be handled properly in future research.

\label{sec:conclusion}
Based on these observations, we suggest that in addition to label noise, more attention be paid to the shifted label distribution in distantly supervised relation extraction research. We hope that the analysis presented will provide new insights into this long-overlooked factor and encourage future research of creating models robust to label distribution shift. We also hope that methods such as threshold techniques and bias adjustment become useful tools in future research.

\section*{Acknowledgments}
This work has been supported in part by National Science Foundation SMA 18-29268, DARPA MCS and GAILA, IARPA BETTER, Schmidt Family Foundation, Amazon Faculty Award, Google Research Award, Snapchat Gift and JP Morgan AI Research Award. We would like to thank all the collaborators in INK research lab for their constructive feedback on the work.

\bibliography{emnlp-ijcnlp-2019}
\bibliographystyle{acl_natbib}

\clearpage

\appendix
\section{Detailed Experiment Setup}
\label{sec:appendix-exp}
For a fair and meaningful comparison, we use the same experimental setup in all experiments.





\subsection{Model Details}
We consider two popular classes of relation extraction methods here, \ie, feature-based and neural models. 
For each relation mention, these models will first construct a representation $\h$, and then make predictions based on $\h$. \footnote{CoType and Logistic Regression are exceptions as they don't adopt softmax to generate output.}
\begin{equation*}
p(y = r_i|\h) = \frac{\exp(\r_i^T\h + b_i)}{\sum_{r_j}\exp(\r_j^T\h+b_j)}
\end{equation*}
where $\r_i$ and $b_i$ are the parameters corresponding to i-th relation type. 

\subsubsection{Feature-based model}
We included three feature-based models, \ie, CoType-RM~\cite{ren2017cotype},  ReHession~\cite{liu2017heterogeneous}, and multi-class logistic regression. For each relation mention $z$, these methods would first extract a list of features, $\F=\{f_1, f_2, ..., f_m\}$.
These features require additional resources like POS-taggers and brown clustering. Detailed description of these features are listed in Table~\ref{tbl:feature}.

\begin{table*}[htbp!]
\centering
\footnotesize
\begin{tabular}{lll}
\hline
\textbf{Feature Name}             & \textbf{Description}                                   & \textbf{Example} \\\hline
Brown cluster            & Brown cluster ID for each token               &  ``\textit{BROWN\_010011001}''       \\
Part-of-speech (POS) tag & POS tags of tokens between two EMs            &  ``\textit{VBD}'' , ``\textit{VBN}'' , ``\textit{IN}''        \\
Entity Mention Token     & Tokens in each entity mention                 &  ``\textit{TKN\_EM1\_Hussein}''        \\
Entity mention (EM) head & Syntactic head token of each entity mention   &  ``\textit{HEAD\_EM1\_HUSSEIN}''       \\
Entity mention order     & whether EM 1 is before EM 2                   &  ``\textit{EM1\_BEFORE\_EM2}''        \\
Entity mention distance  & number of tokens between the two EMs          &  ``\textit{EM\_DISTANCE\_3}''         \\
Entity mention context   & unigrams before and after each EM             &  ``\textit{EM1\_AFTER\_was}''       \\
Tokens Between two EMs   & each token between two EMs                    &  ``\textit{was}'' , ``\textit{born}'' , ``\textit{in}''        \\
Collocations             & Bigrams in left/right 3-word window of each EM &  ``\textit{Hussein was}'' , ``\textit{in Amman}''       \\\hline
\end{tabular}
\caption{Text features used in feature-based models. ("\textit{Hussein}", "\textit{Amman}", "\textit{Hussein was born in Amman}") is used as an example.}
\label{tbl:feature}
\end{table*}

\noindent \textbf{CoType-RM} 
is a variant of CoType~\cite{ren2017cotype}, a unified learning framework to get both the feature embedding and the label embedding. It leverages a partial-label loss to handle the label noise, and uses cosine similarity to conduct inference. 
Here, we only use its relation extraction part.

\noindent \textbf{ReHession} \cite{liu2017heterogeneous}
directly maps each feature to an embedding vector,
treats their average as the relation mention representation $\h$, and uses a softmax to make predictions.
This method was initially proposed for heterogeneous supervision, and is modified to fit our distantly supervised relation extraction task. 
Specifically, for a relation mention annotated with a set of relation $\Y=\{r_{l_1}, \cdots, r_{l_m}\}$, it would first calculate a cross entropy as the loss function:
\begin{equation}
\mathcal{L} = -\sum_{r_i} q(y=r_i |\Y, \h) \log p(y=r_i | \h)
\label{eqn:loss}
\end{equation}
where $p(.| \h)$ is defined in Equation~\ref{eqn:softmax}, and $q(.|\Y, \h)$ is used to encode supervision information in a self-adapted manner:
\begin{equation*}
q(y = r_i|\Y, \h) = \frac{\exp(\r_i^T\h+b_i) \cdot \mathbb{I}(r_i \in \Y) }{\sum_{r_j \in \Y}\exp(\r_j^T\h+b_j)}
\end{equation*}
We can find that when $|\Y| = 1$ (only one label is assigned to the relation mention), $q(.|\Y, \h)$ would be one-hot and Equation~\ref{eqn:loss} would become the classical cross entropy loss. 

\noindent \textbf{Logistic Regression} is applied over the extracted features as a baseline method.\footnote{We use liblinear package from \url{https://github.com/cjlin1/liblinear}}

\subsubsection{Neural Models}
We employed several popular neural structure to calculate the sentence representation $\h$. 
As to the objective function, we use Equation~\ref{eqn:loss} for all the following neural models. 

\noindent \textbf{Bi-LSTMs and Bi-GRUs}
use Bidirectional RNNs to encode sentences and concatenate their final states in the last layer as the representation. 
Following previous work~\cite{zhang2017position}, we use two types of RNNs, Bi-LSTMs and Bi-GRUs. Both of them have 2 layers with 200d hidden state in each layer.

\noindent \textbf{Position-Aware LSTM}
computes sentence representation with an attention over the outputs of LSTMs. 
It treats the last hidden state as the query and integrates a position encoding to calculate the attention~\cite{zhang2017position}.

\noindent \textbf{CNNs and PCNNs}
use convolution neural networks as the encoder. 
In particular, CNN directly appends max-pooling after the convolution layer~\cite{zeng2014relation}; 
PCNN uses entities to split each sentence into three pieces, and does max-pooling respectively on these three pieces after the convolution layer. Their outputs are concatenated as the final output~\cite{zeng2015distant}.


\subsection{Model Training}
We run each model for 5 times and report the average F1 and standard variation.

\noindent\textbf{Optimization.} 
We use Stochastic Gradient Descent (SGD) for all models.
Learning rate is set at 1.0 initially, and is dynamically updated during training, \ie, once the loss on the dev set has no improvement for 3 consecutive epochs, the learning rate will be factored by 0.1.

\noindent\textbf{Hyper-parameters.}
For ReHession, dropout is applied on input features and after average pooling. We tried the two dropout rates in $\{0.0,0.1,0.2,0.3,0.4,0.5\}$.

For Position Aware LSTM, Bi-LSTM and Bi-GRU, dropout \cite{srivastava2014dropout} is applied at the input of the RNN, between RNN layers and after RNN before linear layer. Following \cite{melis2017state} we tried input and output dropout probability in $\{0.4,0.5,0.6,0.7,0.8\}$, and intra-layer dropout probability in $\{0.1,0.2,0.3,0.4\}$.
We consider them as three separate hyper-parameters and tune them greedily.
Following previous work \cite{zhang2017position}, dimension of hidden states are set at 200.

Following previous work~\cite{lin2016neural}, the number of kernels is set to 230 for CNNs or PCNNs, and the window size is set at 3. Dropout is applied after pooling and $\tanh$ activation . We tried the dropout rates in $\{0.1,0.2,0.3,0.4,0.5\}$.

\section{Additional Figures and Tables}
\label{sec:appendix}



\begin{figure*}[htbp!]
    \centering
    \includegraphics[trim={2cm 0cm 2cm 0cm},clip,width=\textwidth]{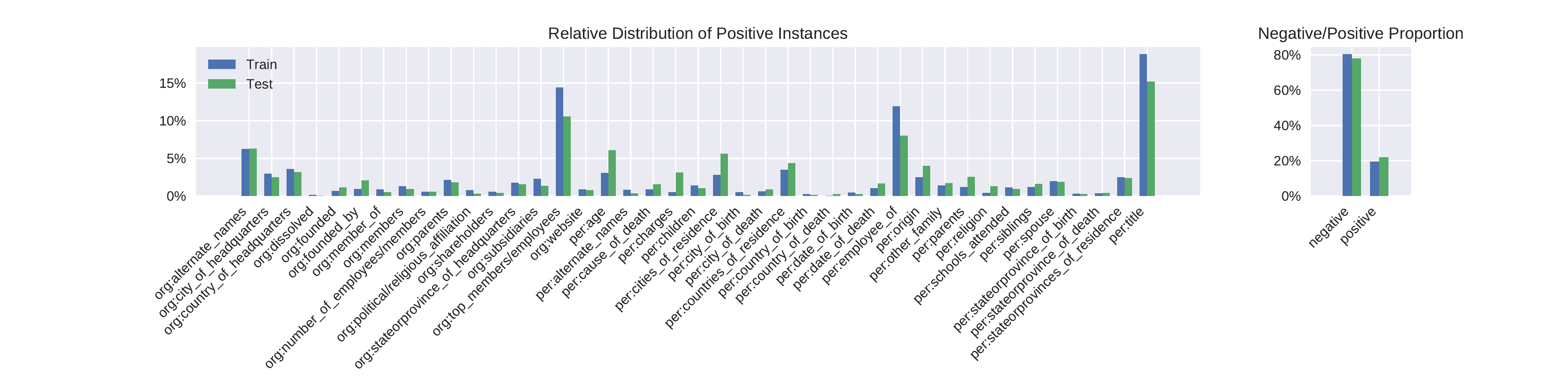}
    \includegraphics[trim={2cm 0cm 2cm 0cm},clip,width=\textwidth]{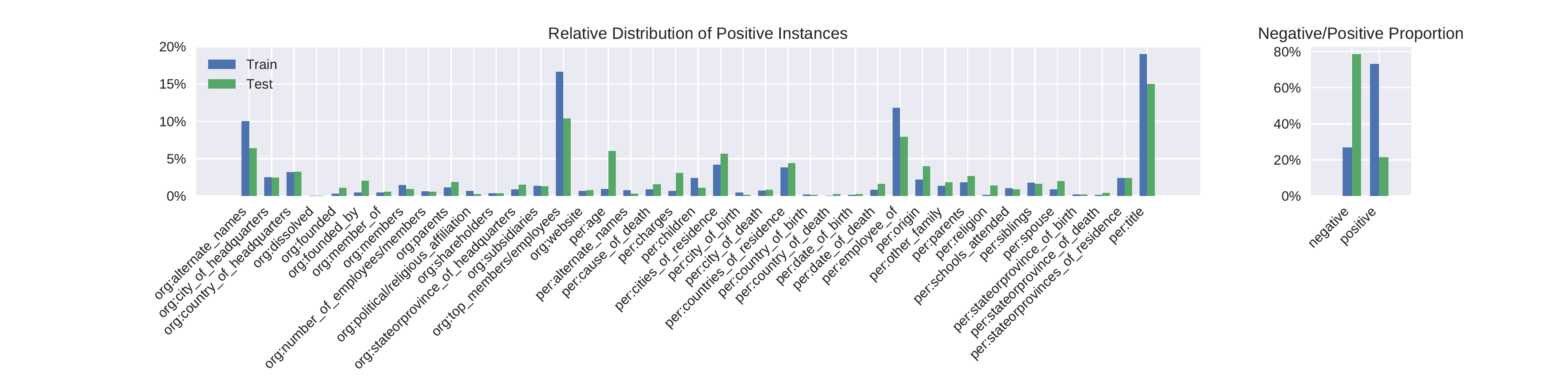}
    \caption{\textbf{Top:} Label distribution of original TACRED; \textbf{Bottom:} Using a randomly generated distribution for $\mathsf{S5}$ train set, and keeping original test set. Label distribution of other synthesized datasets ($\mathsf{S1}$-$\mathsf{S4}$) are generated with linear interpolation of these two train set distributions.}
    \label{fig:TACRED original}
\end{figure*}

\begin{figure*}[htbp!]
    \centering
    \includegraphics[trim={2cm 0cm 2cm 0cm},clip,width=\textwidth]{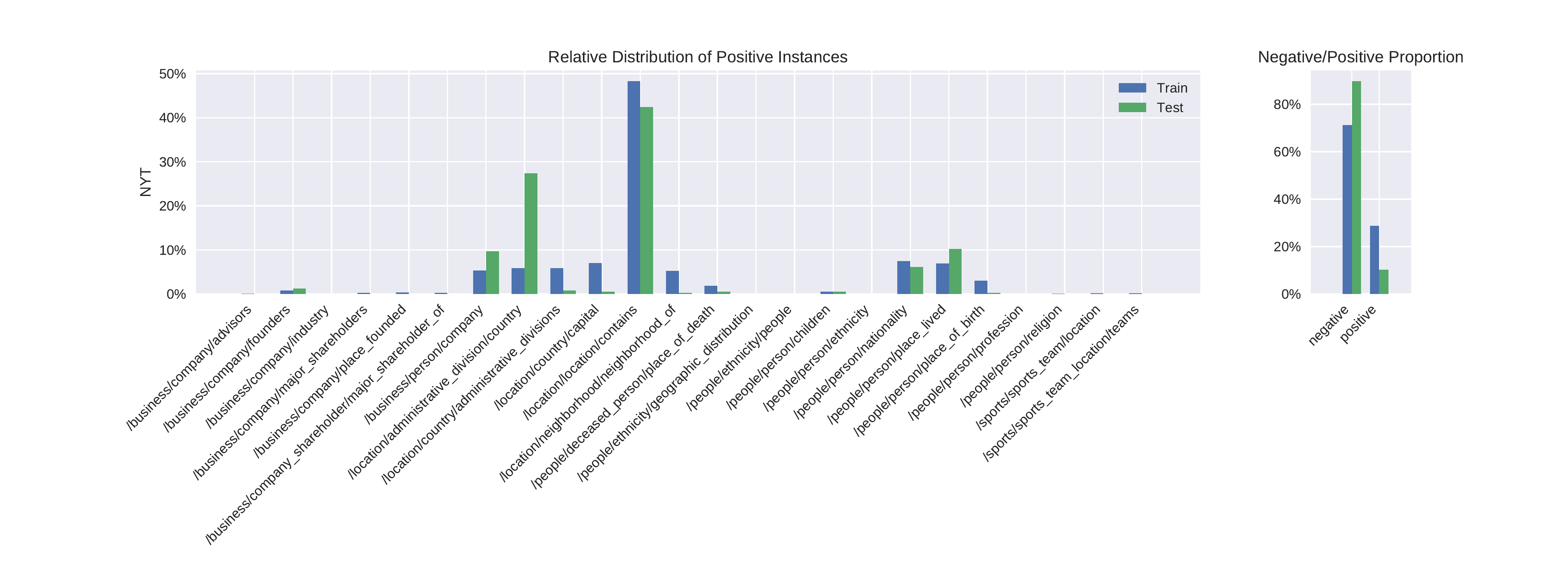}
    \vspace{-0.5cm}
    \caption{Label Distribution of original NYT. Similar to KBP, label distributions are shifted.}
    \label{fig:NYT original}
\end{figure*}

Figure \ref{fig:TACRED original} shows the full label distribution of TACRED dataset, and the simulated TACRED $\mathsf{S5}$ dataset with a shifted distribution. Figure \ref{fig:NYT original} shows the full label distribution of NYT dataset. NYT is constructed with distant supervision and has a shifted distribution.

\end{document}